\documentclass{revtex4}
\usepackage{times,graphicx}
\usepackage{url,amsmath,amssymb,soul,mathpazo}
\usepackage[Alg]{algorithm}
\usepackage{algpseudocode}

\def \LCA {\rm LCA}

\def \Gam {\mathsf{Gam}}
\def \Bet {\mathsf{Beta}}
\def \sv {\mathbf{s}}
\def \zv {\mathbf{z}}

\def \Rb {\mathbb{R}}
\def \lmdv {\boldsymbol{\lambda}}
\def \thev {\boldsymbol{\theta}}
\def \Vol {{\rm vol}}
\newcommand{\Ind}[1]{\mathbb{1}_{\{#1\}}}
\newcommand{\E}[1]{\left\langle #1 \right\rangle}

\begin{document}

\title{Fast and reliable inference algorithm for hierarchical stochastic block models}

\author{Yongjin Park}
\affiliation{Department of Biomedical Engineering, Johns Hopkins University, Baltimore, MD}
\author{Joel S. Bader}
\affiliation{Department of Biomedical Engineering, Johns Hopkins University, Baltimore, MD}

\maketitle

\section{Introduction}

Network clustering reveals the organization of a network or
corresponding complex system with elements represented as vertices and
interactions as edges in a (directed, weighted) graph.  Although the
notion of clustering can be somewhat loose, network clusters or groups
are generally considered as nodes with enriched interactions
\cite{Newman:2001wm,Leskovec:2008im} and edges sharing common patterns
\cite{Ahn:2010dj}.  Statistical inference often treats groups as
latent variables, with observed networks generated from latent group
structure, termed a stochastic block model \cite{Nowicki:2001id}.
Regardless of the definitions, statistical inference can be either
translated to modularity maximization, which is provably an
NP-complete problem \cite{Brandes:2008bu}.

Here we present scalable and reliable algorithms that recover
hierarchical stochastic block models fast and accurately.  Our
algorithm scales almost linearly in number of edges, and inferred
models were more accurate that other scalable methods.

\section{Models}

\paragraph{Stochastic block model}
We represent a network data as an adjacency matrix $A$, where each
element $A_{ij}$ captures the relationship between vertices $i, j \in V$.
Let $n$ be a number of vertices, $n=|V|$, and $m$ be a number of
(un)directed edges, or non-zero entries of (upper triangular) $A$.  For a binary
and undirected network, $A_{ij}$ takes on a value $1$ if $i$ and $j$
interact, $0$ otherwise.  We assume that the set $V$ decomposes into
an arbitrary $K$ disjoint subsets.  We give a unique index $1,
\ldots, K$ to these subsets.  For simplicity let $g_i$ denotes
membership mapping of vertex $i$ to a certain group, with $g_i \in
[K]$.  Assume each group (or block) is identifiable by a $K\times K$
block matrix $\thev$ and $\theta_{ab} \in [0,1]$ for all ${a,b \in [K]
  \times [K]}$.  We can define the data likelihood as
\[
p(A|g,\theta) = \prod_{i<j} \theta_{g_i g_j}^{A_{ij}} (1-\theta_{g_i,g_j})^{1-A_{ij}}.
\]

\paragraph{Degree-corrected stochastic block model}

In some situations, a degree-corrected stochastic block model (DSBM)
provides more appealing group structures in a real-world network
\cite{Karrer:2010vd}.  Again, we represent a network as an
adjacency matrix $A$; a vertex set $V$ decomposes into $K$ disjoint
subsets; the membership function is $g$, ${g:\, V
  \to [K]}$.  However, $A_{ij}$ may takes on values of
non-negative real number, and block-block relationships are captured
by a $K\times K$ matrix $\lmdv$ and $\lambda_{ab} \in \Rb^{+}$ for all
${a,b \in [K] \times [K]}$.

Our DSBM generalizes a regular SBM, weighting edges based on degrees
of endpoints by $\rho_{ij}$ for each $A_{ij}$.  Here we define
$\rho_{ij} \equiv d_{i} d_{j} / \sum_{j'} d_{j'}$.  Note the
denominator determines the scaling of edge weight, but does not 
strongly influence inference results.  We define the data likelihood to have 
each block-block relationship follow the Poisson distribution:
\[
  p(A|g,\lambda) = \prod_{i<j} \frac{(\lambda_{g_i, g_j} \rho_{ij})^{A_{ij}}}{A_{ij}!} e^{-\lambda_{g_i, g_j}\rho_{ij}}.
\]

\paragraph{Hierarchical group-group structure}

Extensive empirical studies such as \cite{Leskovec:2008im} suggest that
the size of a group is limited to a certain number, meaning that ${n/K
  < \infty}$ and ${K \to \infty}$ as ${n \to \infty}$.  Therefore a
full modeling of the block matrices $\lmdv$ and $\thev$ is inherently
formidable \cite{Choi:2012ip} since $K^2$ can easily exceed $n$ and
$m$.  However, most state-of-the-art algorithms require $O(K^2)$
operations per pair and iteration (e.g., \cite{Airoldi:2008wi}), which
essentially scales in $O(n^2 K^2) \approx O(n^4)$.  A model can be
reduced that may include only within-group relationships, e.g.,
diagonalization of the matrix $\thev$ \cite{Gopalan:2012vi}, so that
we achieve $O(mK)$ runtime where $m$ is number of edges.  This
strategy is useful if groups are identifiable by just ``within-group''
edges, and edges between groups are sampled with some background
probability.

\begin{figure}[h!]
  \centering
  \includegraphics[width=.8\textwidth]{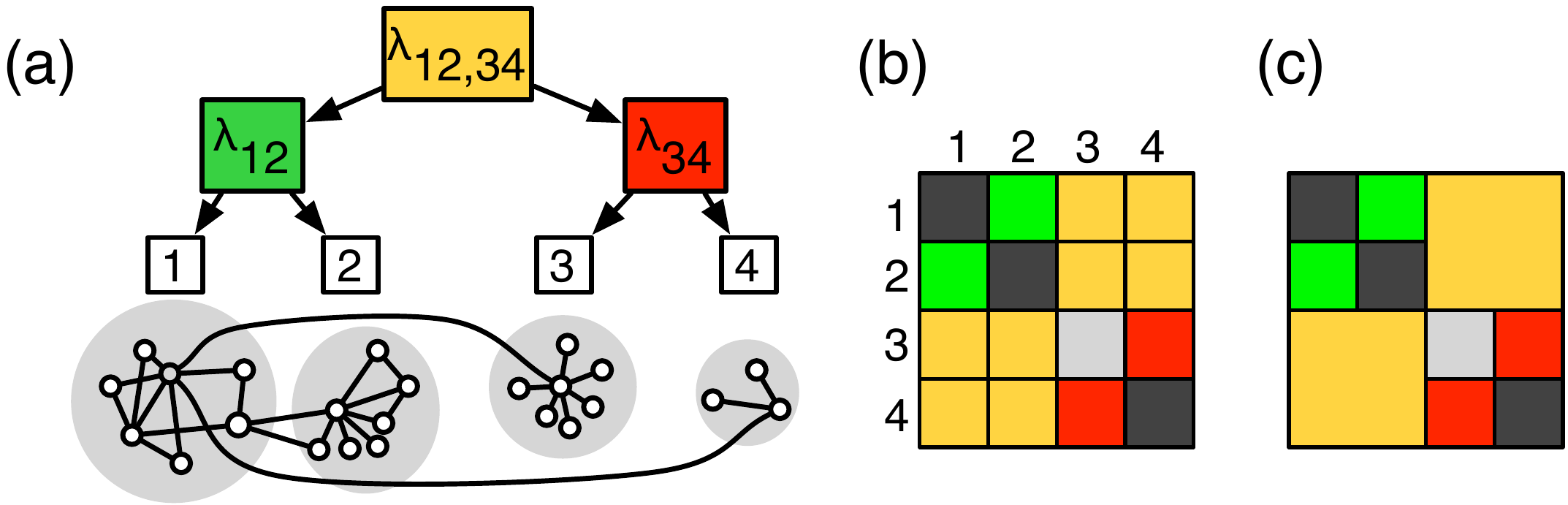}
  \caption{Hierarhical stochastic block model.  (a) We assume that
    group-group relationships are shared hierarchically. (b) A full
    block matrix.  (c) A shrunken block matrix by the hierarchical
    model.}
  \label{fig:h.model}
\end{figure}

Hierarchical assumption provides balance between the full and the
diagonal block-block relationships.  We adopt the idea of
\cite{Clauset:2008fx} and model the hierarchical group-group relations
by a binary dendrogram (see Fig.\ref{fig:h.model}).  For instance, the
relationship between group $a$ and $b$ is captured by lowest common
ancestor $\LCA(a,b)$ in the binary dendrogram and associated
parameters.  The model complexity of hierarchical structure scales in
$O(K)$ while modeling group-group interactions.

For the hierarchical SBM (hSB), the likelihood function is defined
\begin{equation}
  \label{eq:hsb.likelihood}
  p(A|g,\thev) = \prod_{i<j} \theta_{\LCA(g_i, g_j)}^{A_{ij}} (1-\theta_{\LCA(g_i,g_j)})^{1-A_{ij}}
\end{equation}
where the $\theta$ parameters follow the conjugate Beta distribution,
\[
p(\theta|a_0, b_0) = \Bet(\theta|a_{0},b_{0}) = \frac{\Gamma(a_0+b_0)}{\Gamma(a_0)\Gamma(b_0)} \theta^{a_0 -1} (1-\theta)^{b_0 -1}.
\]
For the hierarchical DSBM (hDSB), the likelihood function is defined
\begin{equation}
  \label{eq:hdsb.likelihood}
  p(A|g,\lambda) = \prod_{i<j}\frac{ (\lambda_{\LCA(g_i,g_j)} \rho_{ij})^{A_{ij}}}{A_{ij}!} e^{-\lambda_{\LCA(g_i,g_j)} \rho_{ij}},
\end{equation}
where the rate parameters follow the Gamma distribution,
\[
p(\lambda|a_0, b_0) = \Gam(\lambda|a_{0},b_{0}) = \frac{b_0^{a_0}}{\Gamma(a_0)} \lambda^{a_0 -1} e^{-\lambda b_0}.
\]
% We used $a_0=b_0=1$ for the hSB and $a_0=1$ and $b_0=2$ for the hDSB.

\section{Model inference}

\subsection{Variational Bayes inference}

It is intractable to search over all binary dendrogram models.
Instead, we restrict the model space to a class of complete binary
trees with a fixed yet sufficiently deep depth.  We then apply
variational inference algorithms: mean-field approximation
\cite{Xing:2003ua} and locally collapsed variational inference
\cite{Wang:2012uu}.

We rewrite the likelihood of a tree model $\mathcal{T}$
(Eq.\ref{eq:hdsb.likelihood}) by introducing latent variables, $z_{ia}$
indicating membership of a vertex $i$ in a group $a$ if $z_{ia}=1$,
otherwise $z_{ia}=0$.
\[
p( A| z, \lambda ) = \prod_{i < j} \prod_{r \in \mathcal{T}} \left[ \frac{\lambda_{r} \rho_{ij}}{A_{ij}!} e^{-\lambda_{r} \rho_{ij}} \right]^{z_{ia} z_{jb} \Ind{ \LCA(a,b) = r}}
\]
We approximate the posterior distribution of $z$ and $\lambda$ by the
following surrogate distribution:
\begin{eqnarray}
  \label{eq:var.surrogate}
  q(z,\lambda|\mu,\alpha,\beta) &=& 
\prod_{i=1}^{n} q(\zv_{i}|\mu_{i})
\prod_{r  \in \mathcal{T}} q(\lambda_{r}|\alpha_{r},\beta_{r})
\end{eqnarray}
where
$q(\zv_{i}|\mu_{i}) = \prod_{k=1}^{K} \mu_{ia}^{z_{ia}}$
and 
$q(\lambda_{r}|\alpha_{r}, \beta_{r}) = \Gam(\lambda_{r}| \alpha_{r}, \beta_{r})$.
As in \cite{Wang:2012uu}, we distinguish random variables $\{\zv_{i}\}$ as 
local variables and $\{\lambda_{r}\}$ as global variables.

\paragraph{Global update}
Given $\E{z_{ia}} \equiv \mu_{ia}$, 
by the generalized mean-field theory \cite{Xing:2003ua},
we have variational distributions
$ q(\lambda_{r}| \alpha_{r}, \beta_{r}) $,
for all $r$, characterized by
\begin{eqnarray}
  \alpha_r = a_0 + \sum_{a,b} \sum_{i<j} \Ind{r=\LCA(a,b)} \mu_{ia} \mu_{jb} A_{ij}, \quad
  \beta_r = b_0 + \sum_{a,b} \sum_{i<j} \Ind{r=\LCA(a,b)} \mu_{ia} \mu_{jb} \rho_{ij}. \label{eq:update.lambda}
\end{eqnarray}
We can take expectation with respect to the variational Gamma
distributions, 
${\E{ \lambda_r } = \alpha_r / \beta_r}$
and ${\E{ \ln\lambda_r } = \psi(\alpha_r) - \ln \beta_r}$,
where $\psi(\cdot)$
denotes the digamma function.  We let
\begin{equation}
\eta_{r} \equiv (\ln\lambda_{r}, -\lambda_{r})^{\top}\label{eq:nat.param}
\end{equation}
for convenience.

\paragraph{Local update by mean-field theory}
Let us consider that we update the probability $\mu_{i}$ of assignment
of a vertex $i$ to a certain group $a$, given all other latent
assignments %
${\{\mu_{j}:\, j\neq i\}}$ and the global variables fixed.  Note that
this probability depends on the global variables $\eta_{r}$ located
along the path from the leaf $a$ to the root of the tree model, which
we denote by $\Pi(a)$.

For simplicity, let 
\begin{equation}
  d_{ia} \equiv \sum_{j} A_{ij} \mu_{ja}, \quad
  \sv_{ia} \equiv (d_{ia},\,\sum_j \mu_{ja} \rho_{ij})^{\top}.\label{eq:suff.stat.leaf}
\end{equation}
At some $r \in \Pi(a)$, we may use the aggregated statistics:
\begin{equation}
  \label{eq:suff.stat}
  \sv_{ir} \equiv \sum_{b} \Ind{ r = \LCA(a,b) } \sv_{ib},
\end{equation}
which allows us to write the update equation sufficiently as
\begin{equation}
  \label{eq:update.mu.mf}
  \mu_{ia} \propto \exp\{ 
    \sum_{r \in \Pi(a)}
    \E{\eta_{r}}^{\top} \sv_{ir}
  \}
\end{equation}
subject to $\sum_{a=1}^{K} \mu_{ia} = 1$ and $\mu_{ia} \ge 0$.

\paragraph{Local update by collapsed variational inference}
Alternatively, we may update $\mu_{ia}$ by collapsing the global
variables with respect to the global variational distributions.
Again, we consider assigning a vertex $i$ to group $a$ and use
$\sv_{ir}$ (Eq.\ref{eq:suff.stat}) collected along the path $\Pi$ from
the root of the tree to the group $a$.  However, instead of using 
the expected natural parameters (Eq.\ref{eq:nat.param}), we
integrate them out.
\begin{equation}
  \mu_{ia} 
  \propto \prod_{r \in \Pi(a)} \E{ p(\sv_{ir}|\lambda_{r}) }_{q(\lambda|\alpha_{r},\beta_{r})} 
  \label{eq:update.mu.lcvi}
\end{equation}
where 
\[ 
  \E{ p(\sv_{ir}|\lambda_{r}) }_{q(\lambda|\alpha_{r},\beta_{r})} \propto
  \int \exp\{ \eta_{r}^{\top}\sv_{ir} \} q(\eta_{r}|\alpha_{r}, \beta_{r})\, d\lambda_{r}.
\]
Since our model space has fixed size, we need not sample 
as proposed in \cite{Wang:2012uu}, but just normalize $\mu_{i}$
to satisfy the constraints:
$\sum_k \mu_{ia} = 1$ and $\mu_{ia} \ge 0$.

\subsection{Dynamic programming}

The overall algorithm alternates between global and local updates.
We approximate $q(\lambda_{r}|\alpha_{r},\beta_{r})$ by
Eq. \ref{eq:update.lambda}, then locally update $q(\zv_{i}|\mu_{i})$
by Eq. \ref{eq:update.mu.mf} for meanfield, or \ref{eq:update.mu.lcvi}
for locally collapsed variational inference.

\paragraph{Lazy evaluation of the sufficient statistics}
To have it efficiently evaluate required statistics in the global
update (Eq.\ref{eq:update.lambda}), we calculate the following statistics
at the leaf-level, 
\[
d_{ia} = \sum_{j} A_{ij} \mu_{ja}, \quad
n_{ia} = \mu_{ia}, \quad
\Vol_{a} = \sum_{i} \mu_{ia} d_{i},
\]
then, for internal $r$, we can easily accumulate them using
lower-level results,
\[
d_{ir} = d_{i{\rm left}(r)} + d_{i{\rm right}(r)}, \quad
n_{ir} = n_{i{\rm left}(r)} + n_{i{\rm right}(r)}, \quad
\Vol_{r} = \Vol_{{\rm left}(r)} + \Vol_{{\rm right}(r)}.
\]
We can rewrite the update equations (Eq.\ref{eq:update.lambda}) with
respect to $d, n, \Vol$ as follows.
At the leaf group $a$, 
\[
\alpha_{a} = a_0 + \sum_{i} d_{ia} \mu_{ia},\quad
\beta_{a} = b_0 + \frac{1}{4m} \sum_{i} d_i \mu_{ia} (\Vol_{a} - d_{i} \mu_{ia}) 
\]
and for internal group $r$,
\[
\alpha_{r} = a_0 + \sum_{i} n_{i{\rm left}(r)} d_{i{\rm right}(r)}, \quad
\beta_{r} = b_0 + \frac{1}{2m} \sum_{i} d_{i} n_{i{\rm left}(r)} (\Vol_{{\rm right}(r)} - d_i n_{i{\rm right}(r)}).
\]

\paragraph{Deterministic local update}
In practice, a majority of vertices can play a single role, i.e.,
$\E{\zv_{i}}$ has only one non-zero element filled with the value $1$.
We evaluate $\mu_{i}$ by (Eq.\ref{eq:update.mu.mf}) or
(Eq.\ref{eq:update.mu.lcvi}).  Then we set $\mu_{ia^*} = 1$ for $a^{*}
= \arg\max_{a} \sum_{r \in \Pi(a)} \E{\eta_{r}} \sv_{ir}$; $mu_{ia} =
0$ for $a \neq a^{*}$.  

Hierarchical group structure allows us to efficiently search for the
maximum assignments of vertices.  While we perform a depth-first
search over the binary tree model, the algorithm keeps track of
maximum score $M$, corresponding group assignment $k$, and sufficient
statistics $\sv$ (See Alg.\ref{alg:determ.inference} for details).

\begin{algorithm}[h!]
  \begin{algorithmic}
    \Repeat
    \ForAll{$i \in V$}
%    \State $\Vol[k] \gets \Vol[k] - d_i \mu_{ik}$ for all $k$
    \State $d_{ik} \gets \sum_{j}A_{ij}\mu_{jk}$ for all $k$
    \State $(M,k_{\rm max},\sv) \gets$ {\sc maxGradPath}(root of ${\cal T}$)
    \State $\forall k\neq k_{\rm max},\,\mu_{ik} \gets 0$ but $\mu_{ik_{\rm max}} \gets 1$ 
%    \State $n_{k_{\rm max}} \gets n_{k_{\rm max}} + 1$
%    \State $\Vol[k_{\rm max}] \gets \Vol[k_{\rm max}] + d_i$
    \EndFor
    \Until{convergence of $\{n_k\}$}
    \State
    \Function{maxGradPath}{$r$}
    \If {$r$ is leaf-level, cluster $k$}
    \State $\sv \gets (d_{ik}, d_i \Vol[k] / 2m)$
    \State $M \gets \E{ \eta_k}^{\top}\sv$ (Eq.\ref{eq:update.mu.mf}), or $\ln \E{ p(\sv|\lambda) }_{q(\lambda|\alpha_{k},\beta_{k})}$ (Eq.\ref{eq:update.mu.lcvi})
    \State \Return $(M,k,\sv)$
    \Else
    \State $(M_{\rm left}, k_{\rm left}, \sv_{\rm left})\gets$ {\sc maxGradPath}(left($r$))
    \State $(M_{\rm right}, k_{\rm right}, \sv_{\rm right})\gets$ {\sc maxGradPath}(right($r$))
    \State $M_{\rm left} \gets M_{\rm left} + \E{ \eta_r }^{\top} \sv_{\rm right}$
    (Eq.\ref{eq:update.mu.mf}), 
    or $\ln \E{p(\sv_{\rm right}|\lambda)}_{q(\lambda|\alpha_{r},\beta_{r})}$
    (Eq.\ref{eq:update.mu.lcvi})
    \State $M_{\rm right} \gets M_{\rm right} + \E{ \eta_r }^{\top} \sv_{\rm left}$
    (Eq.\ref{eq:update.mu.mf}),
    or $\ln \E{p(\sv_{\rm left}|\lambda)}_{q(\lambda|\alpha_{r},\beta_{r})}$
    (Eq.\ref{eq:update.mu.lcvi})
    \State $M_{\rm max} \gets \max\,\{M_{\rm left}, M_{\rm right}\}$,
    and $k_{\rm max} \gets \arg\max_{k' \in \{{\rm left,right}\}}\, M_{k'}$
    \State \Return $(M_{\rm max}, k_{\rm max}, \sv_{\rm left} + \sv_{\rm right})$
    \EndIf
    \EndFunction
  \end{algorithmic}
  \caption{Deterministic latent variable inference}
  \label{alg:determ.inference}
\end{algorithm}

\paragraph{Probabilistic local update}
The similar lazy-evaluation approach can be applied to the probabilistic
latent variable update for $\mu_{ia}$.  With
memoization of partial log-scores (Eq.\ref{eq:update.mu.lcvi} or
Eq.\ref{eq:update.mu.mf}), we can exactly evaluate $\mu$ within two
passes of tree traversal (see Alg.\ref{alg:prob.inference} for
details).

\begin{algorithm}[h!]
  \begin{algorithmic}
    \Repeat
    \ForAll{$i \in [n]$}
%    \State $\Vol[k] \gets \Vol[k] - d_i \mu_{ik}$ for all $k$
    \State $d_{ik} \gets \sum_{j}A_{ij}\mu_{jk}$ for all $k$
%    \State $\nabla[k] \gets 0$ for all $k$
    \State {\sc calcGradPath}(root of ${\cal T}$)
    \State {\sc sumGradPath}(root of ${\cal T}$, 0)
    \State $\mu_{ik} \propto \exp\{ \nabla[k] \}$
%    \State $n_k \gets n_k + \mu_{ik}$ for all $k$
%    \State $\Vol[k] \gets \Vol[k] + d_i \mu_{ik}$ for all $k$
    \EndFor
    \Until{convergence}
    \State
    \Function{calcGradPath}{$r$}
    \If {$r$ is leaf-level, cluster $k$}
    \State $\sv \gets (d_{ik}, d_i \Vol[k]/2m)$
    \State \Return $\sv$
    \Else
    \State $\sv_{\rm left}\gets$ {\sc calcGradPath}(left($r$))
    \State $\sv_{\rm right}\gets$ {\sc calcGradPath}(right($r$))
    \State $\delta_{\rm left}[r] \gets \E{ \eta_r }^{\top} \sv_{\rm right}$
    (Eq.\ref{eq:update.mu.mf}), 
    or $\ln \E{ p(\sv_{\rm right}|\lambda) }_{q(\lambda|\alpha_{r},\beta_{r})}$ 
    (Eq.\ref{eq:update.mu.lcvi}) 
    \State $\delta_{\rm right}[r] \gets \E{ \eta_r }^{\top} \sv_{\rm left}$ 
    (Eq.\ref{eq:update.mu.mf}), 
    or $\ln \E{ p(\sv_{\rm left}|\lambda) }_{q(\lambda|\alpha_{r},\beta_{r})}$ 
    (Eq.\ref{eq:update.mu.lcvi}) 
    \State \Return $(\sv_{\rm left} + \sv_{\rm right})$
    \EndIf
    \EndFunction
    \State
    \Function{sumGradPath}{$r,\alpha$}
    \If {$r$ is bottom-level, cluster $k$}
    \State $\sv \gets (d_{ik}, n_{k})$
    \State $\nabla[k] \gets \alpha + \E{ \eta_k }^{\top} \sv$
    (Eq.\ref{eq:update.mu.mf}), 
    or $\ln \E{ p(\sv|\lambda) }_{q(\lambda|\alpha_{k},\beta_{k})}$ 
    (Eq.\ref{eq:update.mu.lcvi}) 
    \Else
    \State {\sc sumGradPath}(left($r$), $\alpha+\delta_{\rm left}$[r])
    \State {\sc sumGradPath}(right($r$), $\alpha+\delta_{\rm right}$[r])
    \EndIf
    \EndFunction
  \end{algorithmic}
  \caption{Recursive latent variable inference}
  \label{alg:prob.inference}
\end{algorithm}

Both global and local update steps scale in $O(mK)$.  The
deterministic update converges faster (in $5$ to $10$ iterations),
while the probabilistic one converges in tens of iterations. 

\paragraph{Pruning unnecessarily branching subtrees}
Allowing sufficient depth of the tree model, we may generate an
over-complicated model fitted to noisy observation.  To reduce model
complexity, we apply final pruning steps.  At each subtree of the full
model, we compared this subtree with the collapsed model under the
single group.  We determine whether to collapse or not via the Bayse
factor, or log-ratio of the marginal likelihood
\cite{Heller:2005el}.  This automatically determines the number of
groups from the data.

\paragraph{Initialization}
Variational inference algorithms not necessarily guarantee the
convergence to global optima.  To avoid bad local optima, we may
restart the algorithm multiple times from random configuration.
However, the model space grows super-exponentially and an algorithm
may require exponentially many random restarts.  Instead, we found
that iterative bisections of network provide a good starting point.
Since each bisection using the deterministic inference with $2$ groups
can be completed in $O(m)$, we can finish the whole initialization in
$O(m \log K)$.

\paragraph{Approximation for speed up}
Although $O(m K)$ runtime is practical for small $K$, a network of
10,000 nodes and 100,000 edges could have $K$ as large as 1000.
Therefore, full computation of each variational update could make the
overall algorithm scales essentially in $O(mn)$ (if $O(n)=O(K)$).  
We may reduce $m$ as in the previous work \cite{Gopalan:2012vi} by
stochastic variational inference.  
Here, we address different aspect, reducing $K$ to some $k^* \ll K$.

Suppose we want to re-assign a vertex $i$ by evaluating
${\{\mu_{ik}:\, k \in [K] \}}$.  In assortative networks, vertices
tend to form a group only with connected vertices.  This allows us to
locally carry out the computation of latent and global updates.  Let
$U$ be a subset of leaf groups, to which the vertex $i$ is connected,
i.e., $U=\{k \in [K]: d_{ik} > 0\}$ (see Eq.\ref{eq:suff.stat.leaf}).
Let $k_{\min} = \min U$ and $k_{\max} = \max U$.  With a proper
initial configuration (e.g., iterative bisections), we get $|U| \ll
K$.  Then, we can restrict {\sc maxGradPath}
(Alg.\ref{alg:determ.inference}), and {\sc calcGradPath} and {\sc
  sumGradPath} (Alg.\ref{alg:prob.inference}) to a small subtree.  We
can call them from $\LCA(k_{\min},k_{\max})$.  Moreover, this locality
can be determined in constant time.

\section{Results and Discussions}

\paragraph{Simulation study}
The hDSB outperforms in various benchmark networks.  We generated
sparse network data with average degree $10$ and maximum degree $100$
using the LFR benchmark \cite{Lancichinetti:2008ge}.  We also tested
on benchmark networks with higher degrees (e.g., $30, 50, 100$) and
found that hDSB was still the best performing algorithm, and its
performance can easily attain to the mutual information $1$.  However,
this setting is far from real-world networks, so we omit the result.
With the fixed degree distribution, we varied size of the networks
($5000, 10000$). We also varied the maximum size of groups ($20, 50,
100, 150$) while fixing the minimum size to $10$.  Although our method
can provide mixed membership probability, we did not allow mixed
membership to compare with other community detection methods.

\begin{figure}[h!]
  \centering
  \includegraphics[width=\textwidth]{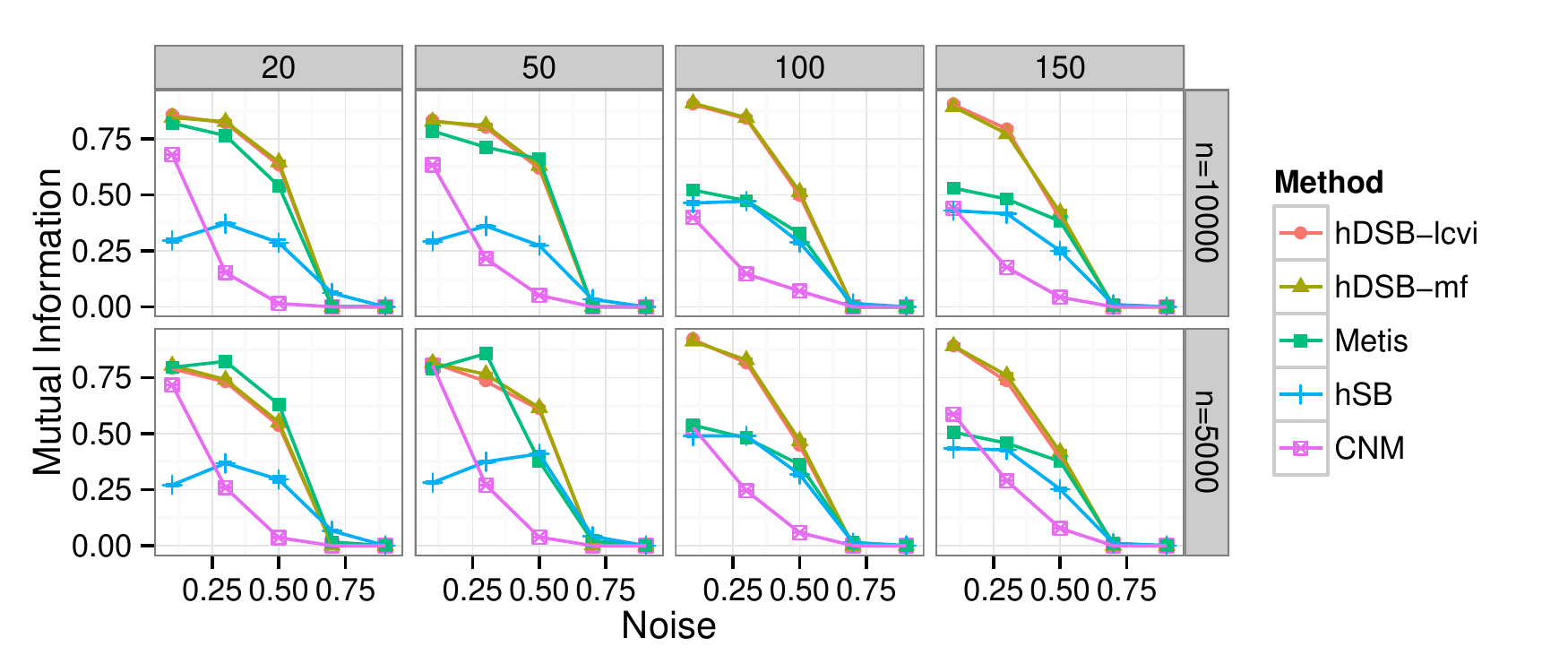}
  \caption{The benchmark result.  {\bf x-axis}: noise parameter (the
    $\mu$ parameter of LFR \cite{Lancichinetti:2008ge}); {\bf y-axis}:
    normalized mutual information \cite{Lancichinetti:2009dy}; {\bf
      Top titles}: the maximum size of a group; {\bf Right titles}:
    total number of vertices. See the text for details.}
  \label{fig:benchmark}
\end{figure}

Fig.\ref{fig:benchmark} compares
hDSB with locally collapsed variational inference (hDSB-lcvi), hDSB
with mean-field (hDSB-mf), $k$-metis provided with true $K$
\cite{Karypis:1998by} (Metis), hSB with mean-field (hSB), and
modularity maximization \cite{Clauset:2004uy} (CNM).  We measured the
normalized mutual information (NMI) \cite{Lancichinetti:2009dy}
between the inferred and the ground truth groups.  The NMI scales in
between $0$ and $1$, and higher value means higher similarity.  

In overall, the performance of hDSB dominates the other methods.  The
effect of different latent variable inference was not so significant.
For networks with balanced group structure, with the maximum group
size $\le 50$, Metis algorithm performs nearly as well as hDSB.
However, it requires the number of groups as a parameter, and
real-world networks may contain heterogeneous group structures.  The
degree-correction provides more realistic group structure than regular
SBM; in fact, we found that it prevents from over-segmentation.  Not
surprisingly, since CNM algorithm relies on local greedy steps, it was
most sensitive to noise and resolution limit problem
\cite{Fortunato:2007js} (under-segmentation).

\paragraph{Cross-validation: link prediction tests}

We used 6 real-world datasets (2 biological and 4 social networks):
{\sf NCAA}: NCAA college football network \cite{Girvan:2002ez}; 
{\sf NetSci}: coauthorship network on the network science \cite{Newman:2006uo};
{\sf Hep-Th}: co-authorship network on the High-Energy Theory Archive \cite{Newman:2001wm};
{\sf Cond}: co-authorship network on the Condensed Matter Archive \cite{Newman:2001wm};
{\sf Reactome}: co-reaction network \cite{Croft:2011ga}; 
{\sf CoExp}: Genemania co-expression network \cite{Mostafavi:2008bz}.

With true group structure unknown, performance may be assessed by
link prediction.  We removed links chosen
uniformly from the observed network. For large networks, such as
{\sf Hep-TH}, {\sf Cond}, {\sf CoExp}, we chose the same number of
non-links.  For smaller networks, {\sf NCAA}, {\sf NetSci}, {\sf
  Reactome}, we chose non-links while preserving the original ratio of
links to non-links.  We preprocessed the networks by iteratively
removing vertices with degree $\le 2$, since these vertices do not
form a group.

In addition to the methods used in the benchmark study, we considered
the sampling method of hierarchical random graph \cite{Clauset:2008fx}
(HRG) and the link community optimization by the maximum likelihood
\cite{Ball:2011wz} (LC).  
These methods do not provide a fixed group structure, rather scores
$s_{ij}$ for a pair $ij$, and a high scoring pair means a possible
link.  For the algorithms that provide a group structure, we estimated
the score $s_{ij}$ by group-wise frequency, 
\[
\hat{\theta}_{ab} =
\frac{\textrm{\# links observed between one end in group $a$ and the other in
group $b$}}
{\textrm{\# possible links between groups $a$ and $b$}}.
\]
For instance, a pair $ij$ takes $\hat{\theta}_{ab}$ as a score if $i$
and $j$ belong to groups $a$ and $b$ respectively.  We allowed 10
random restarts for LC method.  Metis and LC require pre-specified
number of blocks or colors.  We performed discrete grid search over
the parametric space and reported the best results.  As a summary
statistic, we estimated the area under the precision-recall curve
(AUPRC) \cite{Davis:2006kr}.  AUPRC scales in $[0, 1]$ on the small
datasets ({\sf NCAA}, {\sf NetSci}, and {\sf Reactome}); $[0.5, 1]$ on
the large networks ({\sf Hep-Th}, {\sf Cond2005}, {\sf CoExp}).

\begin{figure}[h!]
  \centering
  \includegraphics[width=\textwidth]{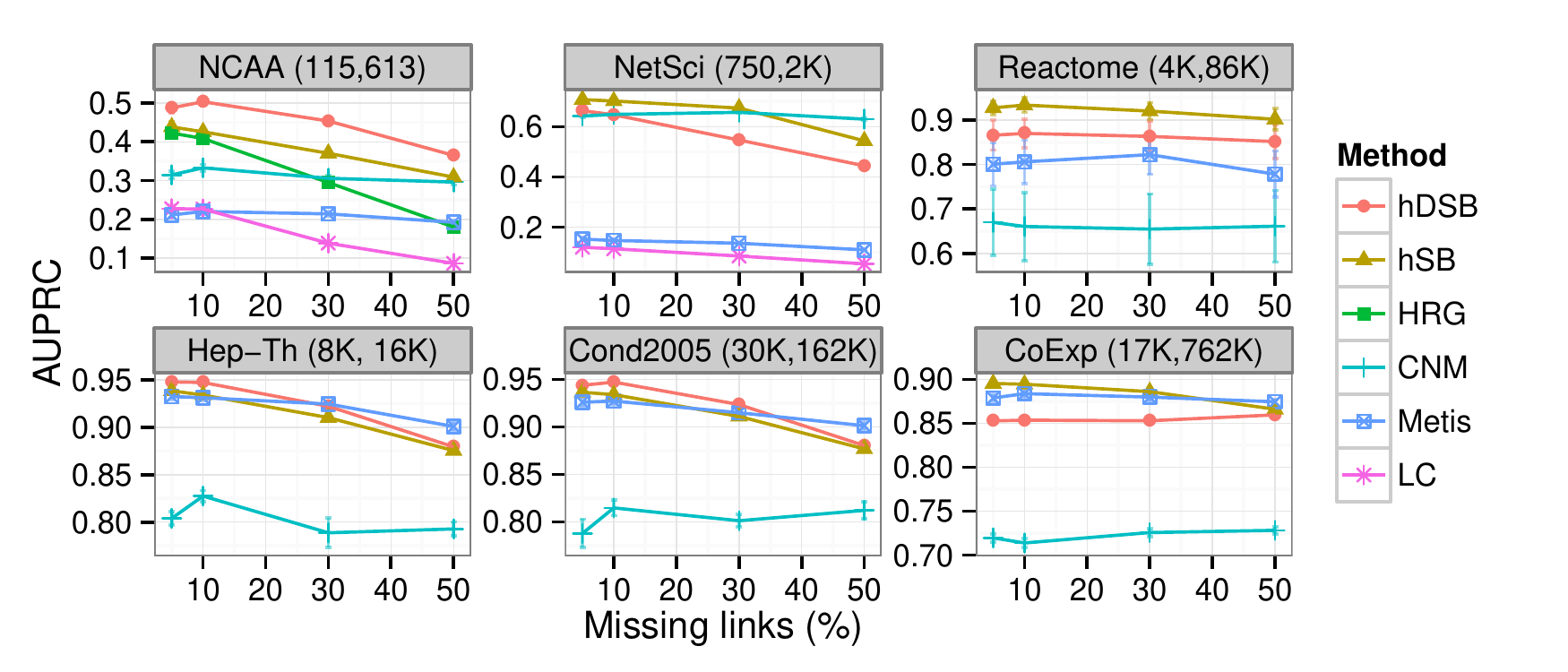}
  \caption{The missing link prediction results. {\bf x-axis}: a
    fraction of missing links; {\bf y-axis}: area under the precision
    recall curve; {\bf (n,m)}: $n=$ number of vertices and $m=$
    number of edges.}
  \label{fig:cv}
\end{figure}

Fig.\ref{fig:cv} shows the results on these networks.
Either hDSB or hSB was consistently top-ranked on all datasets over all
range of missing links.  Notably hSB performed best in some networks.
We may conclude that groups of hSB are as robust as hDSB, and groups
of the benchmark networks can be reasonably further decomposed into
smaller groups.  In the NCAA network, our algorithms even outperformed
the exhaustive sampling method (HRG).  Link community optimization
(LC) works poorly in both scalability and accuracy if a network
consists of many groups.  It appears that the proposed algorithm of
Ball \& Newman \cite{Ball:2011wz} can easily overfit the training
data.  A similar argument was also made previously
\cite{Gopalan:2012vi}.

\paragraph{Runtime}

\begin{figure}[h!]
  \centering
  \includegraphics[width=.8\textwidth]{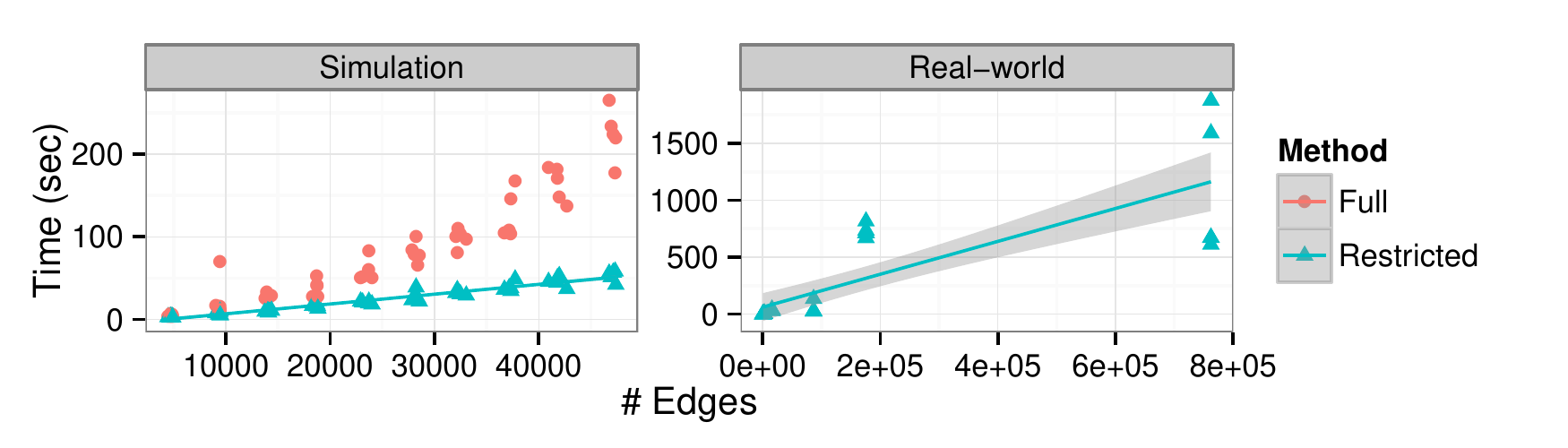}
  \caption{Runtime of the algorithms.}
  \label{fig:runtime}
\end{figure}

Fig.\ref{fig:runtime} shows runtime \footnote{including I/O and
  initialization; measured on Mac Pro 2.8 GHz using a single
  processor.} as a function of number of edges.  We ran
Alg. \ref{alg:prob.inference} with full or restricted calculation.
While the full version increases almost quadratically (red dots),
runtime of the restricted grows linearly (blue triangles) on
simulation and real-world networks.  In terms of accuracy, the
restricted version worked equally well.

\subsubsection*{Acknowledgments}

Source codes available at \url{https://code.google.com/p/hsblock/}
or from the authors by e-mail.

\bibliographystyle{plain}
\bibliography{reference}

\end{document}